# DS4DH at #SMM4H 2023: Zero-Shot Adverse Drug Events Normalization using Sentence Transformers and Reciprocal-Rank Fusion

Anthony Yazdani, Hossein Rouhizadeh, David Vicente Alvarez, Douglas Teodoro
University of Geneva, Geneva, Switzerland

**Abstract**

*This paper outlines the performance evaluation of a system for adverse drug event normalization, developed by the Data Science for Digital Health (DS4DH) group for the Social Media Mining for Health Applications (SMM4H) 2023 shared task 5. Shared task 5 targeted the normalization of adverse drug event mentions in Twitter to standard concepts of the Medical Dictionary for Regulatory Activities terminology. Our system hinges on a two-stage approach: BERT fine-tuning for entity recognition, followed by zero-shot normalization using sentence transformers and reciprocal-rank fusion. The approach yielded a precision of 44.9%, recall of 40.5%, and an F1-score of 42.6%. It outperformed the median performance in shared task 5 by 10% and demonstrated the highest performance among all participants. These results substantiate the effectiveness of our approach and its potential application for adverse drug event normalization in the realm of social media text mining.*

**Introduction**

This paper presents the work of our group - Data Science for Digital Health (DS4DH) - in the Social Media Mining for Health Applications (SMM4H) 2023 task 5. The specific focus of task 5 was the normalization of adverse drug event (ADE) mentions in Twitter to their corresponding preferred terms within the Medical Dictionary for Regulatory Activities (MedDRA) terminology[1]. This normalization task is crucial as it enables the transformation of diverse, heterogeneous textual data into standardized entities, facilitating downstream analysis for pharmacovigilance[2]. Nevertheless, extracting and analyzing health-related information from social media texts is challenging due to the unstructured and informal nature of these data sources. Issues such as slang, abbreviations, misspellings, and informal language create barriers to automated information extraction[3].

In this work, we propose and evaluate a two-stage pipeline for ADE resolution, that is, recognition and normalization of ADE mentions. The first stage involves the use of a BERT[4] fine-tuning approach for named entity recognition[5,6]. The second stage involves a zero-shot normalization via reciprocal-rank fusion[7,8] (RRF) of sentence transformer[9] similarity rankings. This step allows us to associate the identified ADE mentions with their corresponding concepts in MedDRA, irrespective of any previous training on target classes. Our proposed approach stands out due to its simplicity, demonstrating strong performance in comparison with other participants on both in and out of distribution ADEs, while maintaining a straightforward and easily implementable pipeline.

**Methods**

<u>Token classification with BERT</u>: We first fine-tuned BERTweet[10] for ADE token classification. We label each training tweet using the BIOX scheme, where "B" is the beginning of an ADE passage, "I" the inside, "O" the outside, and "X" is used for sub-words, regardless of their association with an ADE. To address the presence of sub-words during classification, we introduced a dummy variable that indicates whether a token is a sub-word (represented as 1) or not (represented as 0). The dummy variables are concatenated to the token embeddings prior to entering the classification layer. The labeled dataset is used to fine-tune BERTweet, which is then used to extract ADE mentions from unseen tweets. In the inference phase, we transform token-based predictions into span-based ADE mentions using a two-step heuristic. Initially, all sub-word tokens (predicted "X") are omitted, transforming a sequence like [B, X, I, I, O, X, I] to [B, I, I, O, I]. Subsequently, any inconsistencies in the sequence that break the continuity of ADE mentions are rectified, converting the previous example further to [B, I, I, I, I].

<u>Normalization using sentence transformers</u>: The second phase deals with the normalization of these mentions using multiple pre-trained sentence transformers. In our final pipeline, we used five pre-trained models: S-PubMedBert[11], all-mpnet-base-v2[9], all-distilroberta-v1[9], all-MiniLML6-v2[9], and a custom sentence transformer wherein we further pre-trained all-MiniLM-L6-v2 on English concepts from the Unified Medical Language System (UMLS)[12]. Specifically, we optimized the all-MiniLM-L6-v2 model for mean squared error over the cosine similarities; that is, English concepts sharing the same unique identifier in UMLS were trained to have a cosine similarity of 1 and negative pairs, generated via negative sampling, were trained to have a cosine similarity of -1.

Linking ADE mentions to MedDRA terms: Each ADE mention detected in a tweet was compared to every MedDRA lowest level terms via cosine similarity of their embeddings, generating a set of ranked similarity scores, i.e., one ranking per ADE per pre-trained transformer. Then, RRF was used to aggregate these ranks into a final rank for each ADE mention. Ultimately, each ADE mention was linked to the MedDRA preferred term with the highest reciprocal-rank.

To identify the best ensemble configuration of pre-trained sentence transformers and the ranking constant of the RRF model, we conducted an extensive grid search. This search covered all possible combinations of our custom model, S-PubMedBert, and all sentence transformers from SBERT[13], as well as the ranking constants within the range of 40 to 60. Notably, our zero-shot approach allowed us to grid-search across all concepts from both the training and validation sets. The final rank was obtained via RRF using the similarity ranks of the five aforementioned sentence transformers with a ranking constant set to 46.

**Results**

As shown in Table 1, our approach demonstrated notable potential, surpassing both the average and median scores in shared task 5 by 10%. Our system achieved the best performances among all the submitting participants with an F1-score of 42.6%, a precision of 44.9%, and a recall of 40.5% overall. When applied to unseen ADEs, our method still managed to exceed the average and median scores, with a relative improvement of 10% upon the median F1-score of the participants. We hypothesize that the reduction in performance for unseen ADEs is likely due to the limited generalizability of the named entity recognition model, given that the normalization employs zero-shot inference and is not constrained by training data. Table 2 shows the performance results from the validation set. We hypothesize that the discrepancy in performance between the validation and test sets may be attributed to the limited sample size present in the validation data.

**Table 1.** Comparative performance of our method against the official mean and median test performances.

| Model | Overall | | | Unseen | | |
|---|---|---|---|---|---|---|
| | **Precision** | **Recall** | **F1-score** | **Precision** | **Recall** | **F1-score** |
| Mean | 0.293 | **0.422** | 0.329 | 0.151 | **0.360** | 0.202 |
| Median | 0.249 | 0.405 | 0.322 | 0.128 | 0.354 | 0.195 |
| Ours | **0.449** | 0.405 | **0.426** | **0.249** | 0.354 | **0.292** |

**Table 2.** Performance of our method on the validation set.

| Model | Overall | | | Unseen | | |
|---|---|---|---|---|---|---|
| | **Precision** | **Recall** | **F1-score** | **Precision** | **Recall** | **F1-score** |
| Ours | 0.500 | 0.447 | 0.472 | 0.100 | 0.400 | 0.160 |

**Discussion and Conclusions**

Our approach significantly outperformed the median score in SMM4H 2023's task 5, surpassing it by a margin of 10% for both overall and unseen ADEs, achieving the best performances among all the submitting participants. This demonstrates its potential for social media text mining in the context of pharmacovigilance. However, diminished performance for unseen ADEs indicates a need to enhance the generalizability of the entity recognition stage. In the broader context of entity resolution, strategies such as incorporating context-aware normalization methods[14], and leveraging large autoregressive language models[15], are promising avenues for improving overall performance and generalizability.